\documentclass[11pt,letterpaper,UTF8]{article}
\usepackage{emnlp2017}
\usepackage{times}
\usepackage{CJKutf8}
\usepackage{bbm}
\usepackage{dsfont}
\usepackage{qtree}
\usepackage{multirow}
\usepackage{url}
\usepackage{latexsym}
\usepackage{graphicx}
\usepackage{tikz-qtree}
\usepackage{mathtools}

\usepackage{amsmath,amssymb,amsthm}

\newcommand{\R}{\ensuremath{\mathbb{R}}}

\newcommand{\si}{\ensuremath{\sigma}}

\newcommand{\ra}{\ensuremath{\rightarrow}}

\newcommand{\myset}[1]{\left\{#1\right\}}
\newcommand{\paren}[1]{\left(#1\right)}

\newcommand{\abs}[1]{\left|#1\right|}

\newcommand{\by}{\ensuremath{\times}}

\newcommand{\norm}[1]{\left|\left|#1\right|\right|}

\emnlpfinalcopy

\def\emnlppaperid{16}

\title{Reconstruction of Word Embeddings from Sub-Word Parameters}

\author{
  {\bf Karl Stratos} \\
  Toyota Technological Institute at Chicago \\
  {\tt stratos@ttic.edu}
  }
\date{}

\begin{document}
\maketitle
\begin{CJK}{UTF8}{gbsn}

\begin{abstract}

Pre-trained word embeddings improve the performance of a neural model at the cost of increasing the model size.
We propose to benefit from this resource without paying the cost by operating strictly at the sub-lexical level.
Our approach is quite simple: before task-specific training, we first optimize sub-word parameters to reconstruct
pre-trained word embeddings using various distance measures.
We report interesting results on a variety of tasks: word similarity, word analogy, and part-of-speech tagging.

\end{abstract}

\section{Introduction}
\label{sec:intro}

Word embeddings trained from a large quantity of unlabled text
are often important for a neural model to reach state-of-the-art performance.
They are shown to improve the accuracy of part-of-speech (POS) tagging from 97.13 to 97.55 \cite{ma2016end},
the F1 score of named-entity recognition (NER) from 83.63 to 90.94 \cite{lample2016neural},
and the UAS of dependency parsing from 93.1 to 93.9 \cite{TACL885}.
On the other hand, the benefit comes at the cost of a bigger model which now stores these embeddings as additional parameters.

In this study, we propose to benefit from this resource without paying the cost by operating strictly at the sub-lexical level.
Specifically, we optimize the character-level parameters of the model to reconstruct the word embeddings prior to task-specific training.
We frame the problem as distance minimization and consider various metrics suitable for different applications,
for example Manhattan distance and negative cosine similarity.

While our approach is simple, the underlying learning problem is a challenging one; the sub-word parameters must reproduce the topology of word embeddings
which are not always morphologically coherent (e.g., the meaning of \texttt{fox} does not follow any common morphological pattern).
Nonetheless, we observe that the model can still learn useful patterns.
We evaluate our approach on a variety of tasks: word similarity, word analogy, and POS tagging.
We report certain, albeit small, improvement on these tasks,
which indicates that the word topology transformation based on pre-training can be beneficial.

\section{Related Work}
\label{sec:related-work}

\newcite{faruqui:2014:NIPS-DLRLW} ``retrofit'' embeddings against semantic lexicons such as PPDB or WordNet.
\newcite{cotterell2016morphological} leverage existing morphological lexicons to incorporate sub-word components.
The aim and scope of our work are clearly different: we are interested in training a strictly sub-lexical model that only operates over characters
(which has the benefit of smaller model size) and yet somehow exploit pre-trained word embeddings in the process.

Our work is also related to \textit{knowledge distillation} which refers to
training a smaller ``student'' network to perform better by learning from a larger ``teacher'' network.
We adopt this terminology and refer to pre-trained word embeddings as the teacher and sub-lexical embeddings as the student.
This problem has mostly been considered for classification and framed as matching the probabilities of the student
to the probabilities of the teacher \cite{ba2014deep,li2014learning,kim2016sequence}.
In contrast, we work directly with representations in Euclidean space.

\section{Reconstruction Method}
\label{sec:reconstruct}

\begin{figure}[t!]
\begin{center}
  \includegraphics[scale=0.55]{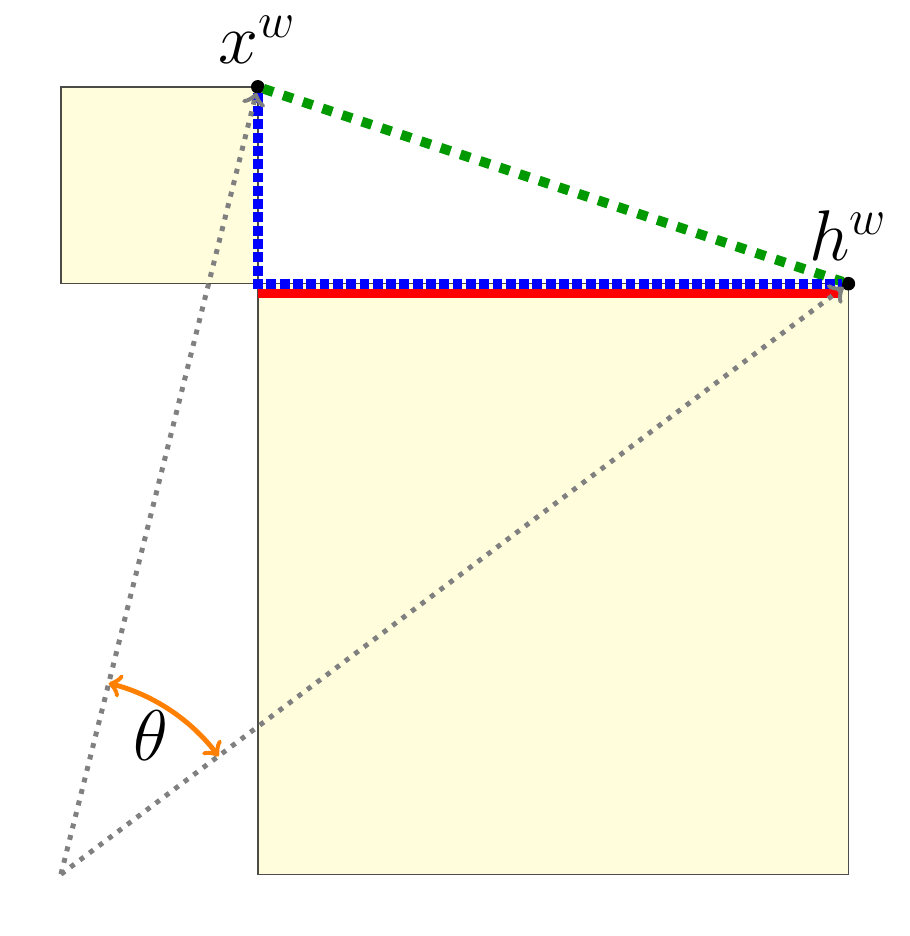}
\caption{Geometric losses corresponding to different distance metrics:
Manhattan distance (blue), Euclidean distance (green), squared error (yellow),
$l_\infty$ distance (red), and negative cosine similarity ($-\cos \theta$).
}
\label{fig:distance}
\end{center}
\end{figure}

Let $\mathcal{W}$ denote the set of word types.
For each word $w \in \mathcal{W}$, we assume a pre-trained word embedding $x^w \in \R^d$
and a representation $h^w \in \R^d$ computed by sub-word model parameters $\Theta$; we defer how to define $h^w$ until later.
The reconstruction error with respect to a distance function $D:\R^d \by \R^d \ra \R$ is given by
\begin{align}
L_D\paren{\Theta} = \sum_{w \in \mathcal{W}} D\paren{x^w, h^w}  \label{eq:reconstruct}
\end{align}
where $x^w$ is constant and $h^w$ is a function of $\Theta$.
Since we use gradient descent to optimize \eqref{eq:reconstruct}, we can define
$D(u, v)$ to be any continuous function measuring the discrepency between $u$ and $v$, for example,
\begin{align*}
D_1(u, v) &:= \sum_{i = 1}^d \abs{u_i - v_i} && \mbox{(Manhattan)} \\
D_{\sqrt{2}}(u, v) &:= \sqrt{\sum_{i = 1}^d \abs{u_i - v_i}^2} && \mbox{(Euclidean)} \\
D_2(u, v) &:= \sum_{i = 1}^d \abs{u_i - v_i}^2 && \mbox{(squared error)} \\
D_\infty(u, v) &:= \max_{i = 1}^d \abs{u_i - v_i} && \mbox{($l_\infty$ distance)} \\
D_{\cos{}}(u, v) &:= \frac{- u^\top v}{\norm{u}_2 \norm{v}_2} && \mbox{(negative cosine)}
\end{align*}
Unlike other common losses used in the neural network literature such as negative log likelihood or the hinge loss,
$L_D$ has a direct geometric interpretation illustrated in Figure~\ref{fig:distance}.
We first optimize \eqref{eq:reconstruct} over sub-word model parameters $\Theta$ for a set number of epochs,
and then proceed to optimize a task-specific loss $L(\Theta, \Theta')$ where $\Theta'$ denotes all other model parameters.

\subsection{Analysis of a Linear Model}

In general, $h^w$ can be a complicated function of $\Theta$.
But we can gain some insight by analyzing the simple case of a linear model, which corresponds to the top layer of a neural network.
More specifically, we assume the form
\begin{align*}
h^w_i = \theta_i^\top z^w &&\forall i = 1 \ldots d
\end{align*}
where $z^w \in \R^{d'}$ is fixed and $\Theta = \{\theta_1 \ldots \theta_d\} \subset \R^{d'}$ is the only parameter to be optimized.

\paragraph{Manhattan distance}
The error $L_{D_1}(\Theta)$ is now
\begin{align*}
L_{D_1}(\Theta) &= \sum_{w \in \mathcal{W}} \sum_{i=1}^d \abs{x^w_i - \theta_i^\top z^w} = \sum_{i=1}^d \mbox{LAD}_i(\theta_i)
\end{align*}
where $\mbox{LAD}_i(\theta) := \sum_{w \in \mathcal{W}} \abs{x^w_i - \theta_i^\top z^w}$ is least absolute deviations (LAD).
It is well-known that the LAD criterion is robust to outliers.
To see this, if $z^w =(1/d') \textbf{1}$ for all $w \in \mathcal{W}$, then a minimizer of $\mbox{LAD}_i(\theta)$ is given analytically by
\begin{align*}
\theta_i^* = \mbox{median} \myset{x^w_i:\; w \in \mathcal{W}}
\end{align*}
where the median resists extreme values (e.g., the median of both $\{1, 2, 3\}$ and $\{1, 2, 999\}$ is $2$).
Thus using Manhattan distance can be useful when teacher word embeddings are noisy
or there are occasional exceptions in morphological patterns that are best ignored.

\paragraph{Squared error}
The error $L_{D_2}(\Theta)$ is now
\begin{align*}
L_{D_2}(\Theta) &= \sum_{w \in \mathcal{W}} \sum_{i=1}^d  \abs{x^{w}_i - \theta_i^\top z^w}^2 = \sum_{i=1}^d  \mbox{OLS}_i(\theta_i)
\end{align*}
where $\mbox{OLS}_i(\theta) := \sum_{w \in \mathcal{W}} \abs{x^w_i - \theta^\top z^w}^2$ is ordinary least squares (OLS).
Thus if the matrix $Z \in \R^{\abs{\mathcal{W}} \by d'}$ with $z^w$ as rows has rank $d'$, the unique solution is given by
$\theta_i^* = \paren{Z^\top Z}^{-1} Z^\top x^w_i$.
Let $\bar{h}^w_i = (\theta_i^*)^\top z^w$ denote the optimal sub-word embedding value.
It is well-known that the change in $\bar{h}^w_i$ caused by removing $x^w_i$ from the dataset is proportional to
the residual $x^w_i - \bar{h}^w_i$ \cite{davidson1993estimation}.
In other words, squared error is sensitive to outliers and may not be as suitable as Manhattan distance for fitting noisy
or incoherent word embeddings.

\paragraph{Other distance metrics}
Euclidean distance is geometrically intuitive but less mathematically convenient than squared error,
thus we choose not to focus on it.
$l_\infty$ distance penalizes the dimension with maximum absolute difference
and can be useful if calculating one coordinate at a time is convenient.
Finally, negative cosine similarity penalizes the angle between embeddings.
This is suitable when we only care about directions and not magnitude,
for instance in word similarity where we measure cosine similarities between word embeddings.

There are distance metrics not discussed here that may be appropriate in certain situations.
For instance, the KL divergence is a natural (assymetric) measure if word embeddings are distributions (e.g., over context words).
More generally, we can consider the wide class of metrics in the Bregman divergence \cite{banerjee2005clustering}.

\section{Sub-Word Architecture}
We now describe how we define word embedding $h^w \in \R^d$ from sub-word parameters.
We use a character-based embedding scheme closely following \newcite{lample2016neural}.
We use an LSTM simply as a mapping $\phi:\R^d \times \R^{d'} \ra \R^{d'}$
that takes an input vector $x$ and a state vector $h$ to output a new state vector $h' = \phi(x, h)$.
See \newcite{hochreiter1997long} for a detailed description.

\subsection{Character Model}
\label{sec:char-units}

Let $\mathcal{C}$ denote the set of character types.
The model parameters $\Theta$ associated with this layer are
\begin{itemize}
\item $e^c \in \R^{d_c}$ for each $c \in \mathcal{C}$
\item Character LSTMs $\phi^{\mathcal{C}}_f, \phi^{\mathcal{C}}_b: \R^{d_c} \times \R^{d_c} \ra \R^{d_c}$
\item $W^f, W^b \in \R^{d \by d_c}$,\; $b^{\mathcal{C}} \in \R^d$
\end{itemize}
Let $w(j) \in \mathcal{C}$ denote the character of $w \in \mathcal{W}$ at position $j$.
The model computes $h^w \in \R^d$ as
\begin{align}
f^{\mathcal{C}}_j &= \phi^{\mathcal{C}}_f\paren{e^{w(j)}, f^{\mathcal{C}}_{j-1}} &&\hspace{-1mm}\forall j = 1 \ldots \abs{w}  \notag\\
b^{\mathcal{C}}_j &= \phi^{\mathcal{C}}_b\paren{e^{w(j)}, b^{\mathcal{C}}_{j+1}} &&\hspace{-1mm}\forall j = \abs{w} \ldots 1 \notag\\
z^w &= W^f f^{\mathcal{C}}_{\abs{w}} + W^b b^{\mathcal{C}}_1 + b^{\mathcal{C}} &&\notag\\
h^w_i &= \max\myset{0, z^w_i}\;\; \forall i = 1 \ldots d && \label{eq:word-char}
\end{align}

We also experiment with a highway network \cite{srivastava2015highway} which has been shown to be beneficial for
image recognition \cite{he2015deep} and language modeling \cite{kim2016character}.
In this case, $\Theta$ includes additional parameters $W^{\mbox{\tiny highway}} \in \R^{d \by d}$ and $b^{\mbox{\tiny highway}} \in \R^d$.
A new character-level embedding $\tilde{h}^w$ is computed as
\begin{align}
t &= \si\paren{W^{\mbox{\tiny highway}} h^w + b^{\mbox{\tiny highway}}} \notag \\
\tilde{h}^w &= t \odot h^w + (\textbf{1} - t) \odot z^w \label{eq:word-char-high}
\end{align}
where $\si(\cdot) \in [0,1]$ denotes an element-wise sigmoid function and $\odot$ the element-wise multiplication.
This allows the network to skip nonlinearity by making $t_i$ close to $0$.
We find that the additional highway network is beneficial in certain cases.
We will use either \eqref{eq:word-char} or \eqref{eq:word-char-high} in our experiments depending on the task.

\section{Experiments}
\label{sec:experiments}

\paragraph{Implementation} We implement our models using the DyNet library.
We use the Adam optimizer \cite{kingma2014adam} and apply dropout at all LSTM layers \cite{hinton2012improving}.
For POS tagging and parsing, we perform a $5 \by 5$ grid search over learning rates $0.0001 \ldots 0.0005$
and dropout rates $0.1 \ldots 0.5$ and choose the configuration that gives the best performance on the dev set.
We use the highway network \eqref{eq:word-char-high} for word analogy and parsing and \eqref{eq:word-char} for others.
Note that the character embedding dimension $d_c$ must match the dimension of the pre-trained word embeddings.

\paragraph{Teacher Word Embeddings}
We use 100-dimensional word embeddings identical to those used by \newcite{dyer2015transition}
which are computed with a variant of the skip $n$-gram model \cite{ling2015not}.
These embeddings have been shown to be effective in various tasks \cite{dyer2015transition,lample2016neural}.

\subsection{Word Similarity and Analogy}

\paragraph{Data}
For word similarity, we use three public datasets WordSim-353, MEN, and Stanford Rare Word.
Each contains 353, 3000, and 2034 word pairs annotated with similarity scores.
The evaluation is conducted by computing the cosine of the angle $\theta$ between each word pair $(w_1, w_2)$ under the model \eqref{eq:word-char}:
\begin{align}
\cos(\theta) = \frac{\paren{h^{w_1}}^\top h^{w_2}}{\norm{h^{w_1}}_2 \norm{h^{w_2}}_2} \label{eq:sim-cos}
\end{align}
and computing the Spearman's correlation coefficient with the human scores.
We report the average correlation across these datasets.
For word analogy, we use the 8000 syntactic analogy questions from the dataset of \newcite{mikolov2013linguistic}
and 8869 semantic analogy questions from the dataset of \newcite{mikolov2013efficient}.
We use the multiplicative technique of \newcite{levy2014linguistic} for answering analogy questions.

\begin{table}[t!]
\begin{center}
{
\begin{tabular}{|l|c|c|c|c|c|c|c|c|}
\hline
                &  \multicolumn{5}{c|}{number of reconstruction epochs}        \\
\hline
metric          &   0           &  10             &  20           &  30    &  50  \\
\hline
$D_1$           &  0.03         &  0.09           & 0.11          & 0.12   &  0.13 \\
$D_2$           &  0.03         &  0.12           & 0.12          & 0.14   &  \textbf{0.15} \\
$D_\infty$       &  0.03         &  0.12           & 0.10          & 0.09   &  0.10 \\
$D_{\cos{}}$     &  0.03         &  \textbf{0.13}  & \textbf{0.15} & \textbf{0.15}  &  \textbf{0.15} \\
\hline
\end{tabular}
\caption{Effect of reconstruction on word similarity: the teacher word embeddings obtain score 0.50.}
\label{tab:similarity}
}
\end{center}
\vspace{-1mm}
\end{table}

\paragraph{Result}
Table~\ref{tab:similarity} shows word similarity scores for different numbers of reconstruction training epochs.
The teacher word embeddings obtain 0.5. The sub-word model improves performance from the initial score of 0.03 up to 0.16.
In particular, the negative cosine distance metric which directly optimizes the relevant quantity \eqref{eq:sim-cos}
is consistently best performing.

Table~\ref{tab:analogy} shows the accuracy on the syntactic and semantic analogy datasets.
An interesting finding in our experiment is that for syntactic analogy,
\textbf{a randomly initialized character-based model outperforms the pre-trained embeddings}
and thus reconstruction only decreases the performance.
We suspect that this is because much of the syntactic regularities is already captured by the architecture.
Many questions involves only simplistic transformation, for instance adding \texttt{r} in $\texttt{wise}:\texttt{wiser}\sim\texttt{free}:x$.
The model correctly answers such questions simply by following its architecture, though it is unable to
answer less regular questions (e.g., $\texttt{see}:\texttt{saw}\sim\texttt{keep}:x$).

Semantic analogy questions have no such morphological regularities (e.g., $\texttt{Athens}:\texttt{Greece}\sim\texttt{Havana}:x$)
and are challenging to sub-lexical models. Nonetheless, the model is able to make a minor improvement in accuracy.

\begin{table}[t!]
\begin{center}
{
\begin{tabular}{|l|l|l|}
\hline
Embedding                           &   Syntactic     &  Semantic \\
\hline
random                             &  \textbf{65.21} &  1.13     \\
$D_1$                              &   26.32         &  2.20     \\
$D_2$                              &   27.56         &  \textbf{2.47}     \\
$D_\infty$                          &   45.68         &  1.74     \\
$D_{\cos{}}$                        &   23.77         &  2.22     \\
\hline
teacher                            &   57.42         &  59.58    \\
\hline
\end{tabular}
\caption{Effect of reconstruction on word analogy (10 reconstruction epochs).}
\label{tab:analogy}
}
\end{center}
\vspace{-1mm}
\end{table}

\begin{table}[t!]
\begin{center}
{
\begin{tabular}{|l|l|l|}
\hline
model                          &   accuracy  & lookup \\
\hline
\textsc{full}                  &   97.20     & 43211  \\
\textsc{full+emb}              &   97.32     & 252365 \\
\hline
\textsc{char}                  &   96.93     &  80   \\
\textsc{char($D_1$)}           &   \textbf{97.17}  &  93        \\
\textsc{char($D_2$)}           &   97.08    & 93      \\
\textsc{char($D_{\infty}$)}     &   97.06     & 93     \\
\textsc{char($D_{\cos}$)}       &   97.08     & 93     \\
\hline
\end{tabular}
\caption{POS tagging accuracy with different definitions of $v^w$ (see the main text).
The final column shows the number of lookup parameters.}
\label{tab:pos}
}
\end{center}
\vspace{-1mm}
\end{table}

\begin{table*}[t!]
  \begin{center}
    {\scriptsize
      \begin{tabular}{|l|lllllll|}
        \hline
        \texttt{beautiful} &  wonderful  &prettiest   &gorgeous   &smartest   &jolly      &famous     &sensual       \\
        &  baleful    &bagful      &basketful  &bountiful  &boastful   &bashful    &behavioural   \\
        &  bountiful  &peaceful    &disdainful &perpetual  &primaeval  &successul  &purposeful     \\
        \hline
        \texttt{amazing}   &  incredible &wonderful   &remarkable &terrific   &marvellous &astonishing &unbelievable \\
        &  awaking    &arming      &aging      &awakening  &angling    &agonizing   &among        \\
        &  arousing   &amusing     &awarding   &applauding &allaying   &awaking     &assaying     \\
        \hline
        \texttt{Springfield} &  Glendale & Kennesaw   &Gainesville   &Lynchburg  &Youngstown    &Kutztown &Harrisburg \\
        &  Spanish-ruled  & Serbian-held &Serbian-led  &Spangled &Serbian-controlled   &Schofield    &Sharif-led \\
        &  Stubblefield  & Smithfield   &Stansfield    &Butterfield &Littlefield  &Bitterfeld      &Sinfield   \\
        \hline
      \end{tabular}
      \caption{Nearest neighbor examples: for each word, the three rows respectively show its nearest neighbors using
        pre-trained word embeddings, student embeddings at random initialization \eqref{eq:word-char-high}, and student embeddings optimized for 10 epochs using $D_1$.}
        \label{tab:topology}
    }
      \end{center}
  \end{table*}

\subsection{POS Tagging}
\label{subsec:experiments-english}

We perform POS tagging on the Penn WSJ treebank with 45 tags
using a BiLSTM model described in \newcite{lample2016neural}.
Given a vector sequence $(v^{w_1} \ldots v^{w_n})$ corresponding to a sentence $(w_1 \ldots w_n) \in \mathcal{W}^n$, the BiLSTM model
produces feature vectors $(h_1 \ldots h_n)$.
We adhere to the simplest approach of making a local prediction at each position $i$
by a feedforward network on $h_i$,
\begin{align*}
p(t_i|h_i) \propto \exp\paren{W^2 f(W^1 h_i + b^1) + b^2}
\end{align*}
where $f_i(v) = \max\myset{0, v_i}$ and $W^1, W^2, b^1, b^2$ are additional parameters.
The model is trained by optimizing log likelihood.
We consider the following choices of $v^w$:
\begin{itemize}
\item \textsc{full}: $v^w = e^w \oplus h^w$ uses both word-level lookup parameter $e^w$ and character-level embedding $h^w$ \eqref{eq:word-char}.
\item \textsc{full+emb}: Same as \textsc{full} but the lookup parameters $e^w$ are initialized with pre-trained word embeddings.
\item \textsc{char}: $v^w = h^w$ uses characters only.
\item \textsc{char($D$)}: Same as \textsc{char} but optimized for 10 epochs to reconstruct pre-trained word embeddings with distance metric $D$.
\end{itemize}

Table~\ref{tab:pos} shows the accuracy of these models.
We see that pre-trained word embeddings boost the performance of \textsc{full} from 97.20 to 97.32.
When we use the strictly character-based model \textsc{char} without reconstruction, the performance drops to 96.93.
But with reconstruction, the model recovers some of the lost accuracy.
In particular, reconstructing with the Manhattan distance metric gives the largeset improvement and yields 97.17.

\subsection{Analysis of Student Embeddings}
\label{subsec:analysis}

Table~\ref{tab:topology} shows examples of nearest neighbors.
For each example, the first row corresponds to the teacher, the second to the student \eqref{eq:word-char-high} at random initialization,
and the third to the student optimized for 10 epochs using $D_1$.
The student embeddings at random initialization are already capable of capturing morphological regularities such as \texttt{-ful} and \texttt{-ing}.
With reconstruction, there is a subtle change in the topology.
For instance, the nearest neighbors of \texttt{beautiful} change from \texttt{baleful} and \texttt{bagful} to \texttt{bountiful} and \texttt{peaceful}.
For \texttt{Springfield}, nearest neighbors change from unrelated words such as \texttt{Spanish-ruled} to fellow nouns such as \texttt{Stubblefield}.

\section{Conclusion}
\label{sec:conclusion}

We have presented a simple method for a sub-lexical model to leverage pre-trained word embeddings.
We have shown that by recontructing the embeddings before task-specific training, the model can improve over random initialization on a variety of tasks.
The reconstruction task is a challenging learning problem; while our model learns useful patterns, it is far from perfect.
An important future direction is to improve reconstruction with other choices of architecture.

\bibliographystyle{emnlp_natbib}
\bibliography{reconstruct}

\clearpage\end{CJK}
\end{document}